# An Ontological approach to the construction of problem-solving models

Sabine BRUAUX[a], Gilles KASSEL[b], Gilles MOREL[c]

a, b LaRIA, University of Picardie Jules Verne: {sabine.bruaux, gilles.kassel}@u-picardie.fr

c, CETMEF, Ministry of Equipment and Environment: Gilles.Morel@equipment.gouv.fr

# An ontological approach to the construction of problem-solving models

**Sabine Bruaux**
LaRIA, University of Picardie
Jules Verne
Amiens, France
sabine.bruaux@u-picardie.fr

**Gilles Kassel**
LaRIA, University of Picardie
Jules Verne
Amiens, France
gilles.kassel@u-picardie.fr

**Gilles Morel**
CETMEF, Ministry of Equipment
and Environment
Compiègne, France
Gilles.Morel@equipement.gouv.fr

## ABSTRACT

Our ongoing work aims at defining an ontology-centered approach for building expertise models for the Common-KADS methodology. This approach (which we have named "OntoKADS") is founded on a core problem-solving ontology which distinguishes between two conceptualization levels: at an object level, a set of concepts enable us to define classes of problem-solving situations, and at a meta level, a set of meta-concepts represent modeling primitives. In this article, our presentation of OntoKADS will focus on the core ontology and, in particular, on *roles* - the primitive situated at the interface between domain knowledge and reasoning, and whose ontological status is still much debated. We first propose a coherent, global, ontological framework which enables us to account for this primitive. We then show how this novel characterization of the primitive allows definition of new rules for the construction of expertise models.

## Categories and Subject Descriptors

I.2.4 [Artificial Intelligence]: Knowledge Representation Formalisms and Methods

## Keywords

Knowledge Engineering and Modeling methodologies, Problem-solving models, CommonKADS, Ontology Engineering, Foundational ontologies, DOLCE, Core problem-solving ontologies, OntoKADS, Ontologies of mental objects, I&DA, COM

## INTRODUCTION

Since the late 1990s, the construction of explicit ontologies has been considered as a promising way of improving the knowledge engineering process: the elaboration of domain, task and method ontologies early on in the design of problem-solving models was recommended [22]. At the same time, however, certain components of these problem-solving models - notably *roles* - appeared to have been excluded from ontological treatment [23]. This component (referred to as a *Knowledge role* in the CommonKADS method [21] and situated at the interface between domain knowledge and reasoning) fulfils an important function: it must allow problem-solving methods to be specified in terms which are independent of particular application domains, thus facilitating re-use of these *generic* methods. Even today, the extra-ontological status of this component does not seem to have progressed. In fact, methods which recommend the use of ontologies all resort to a syntactic ploy - *transformation rules* in PROTÉGÉ and *bridges* in UPML - to link domain knowledge and reasoning [6].

Here, we re-examine this presupposition. We show that recent progress in the field of formal ontologies enables one to account for this component in semantic terms, within a coherent ontological framework.

In our previous work [14], we suggested drawing a distinction between two types of role: roles played by *objects* (e.g. *Physician*, *Student*) and those played by *concepts* (e.g. *Hypothesis*, *Sign*). After having likened the latter to CommonKADS' *Knowledge roles*, we gave them the status of a *meta-property*, i.e. along the same lines as Guarino's proposal (making the *role* concept appear in an ontology of universals [12]). However, in 1999, we were not in a position to provide a coherent ontological framework to account for CommonKADS expertise models in their entirety.

Today, we are tackling this issue by using the OntoKADS method [3]. OntoKADS benefits from a broad range of recent work and progress in i) clarifying the notion of *role* in ontological terms [16][18] ; ii) defining (via use of rich axiomatic) a top-level ontology such as DOLCE, the structuring principles of which are explicit [17] ; iii) integrating mental representations (*reified* entities) like Propositions [8] / Descriptions [11] into an ontology; and iv) defining an ontology of *meta-properties* based on a set of clearly identified primitives (*rigidity*, *dependence*, *identity*) [12]. Hence, we now possess an ontological tool-box which is both necessary and sufficient for making this type of proposal.

In the following sections of this article[1], we first present an overview of the OntoKADS method and then focus on its core problem-solving ontology, with particular emphasis on the part of the ontology that deals with *roles*.

---

[1] This article is an extended version of [3].

# OVERVIEW OF OntoKADS

Our proposal consists of a methodology called OntoKADS which, to a great extent, likens the construction of expertise models to the construction of ontologies. The method comprises two main steps (see Figure 1).

In a first step, the knowledge engineer develops a problem-solving-driven application ontology whose concepts are labeled by modeling primitives. To do this, the method uses an ontology (also named OntoKADS) composed of two main sub-ontologies:

- A core problem-solving sub-ontology which enables the engineer to define (by specialization) the application's concepts and specific reasonings. This sub-ontology extends the high-level DOLCE ontology (Descriptive Ontology for Linguistic and Cognitive Engineering) [17][2].

- A meta-level sub-ontology coding the modeling primitives, which allows the engineer to label the concepts from the previous ontology by using meta-properties which represent modeling primitives. This type of practice is analogous to the labeling advocated in the OntoClean method [13][3].

A software module (see Figure 1) then automatically translates this *labeled* ontology into three subcomponents of an expertise model which resembles CommonKADS [21]: a domain model, an inference model and a task model. This translation principally involves the extraction and reorganization of representations.

In a second step, the knowledge engineer further specifies the problem-solving methods linked to the tasks which he/she has identified.

A software environment for running this method is currently being developed (in the Conclusion section, we explain our choice of software tools). In the remainder of this article, our presentation of OntoKADS will concentrate on describing the ontology's content. We shall first show how the OntoKADS core ontology extends the DOLCE high-level ontology.

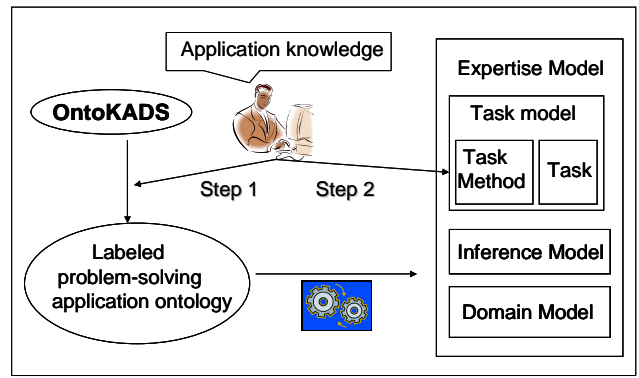

**Figure 1.** Main steps in the OntoKADS method.

# HOW OntoKADS EXTENDS DOLCE

## DOLCE and the notion of *knowledge*

The OntoKADS ontology is defined as an extension of the DOLCE ontology, which means that OntoKADS' concepts and relations are defined by specialization of the abstract concepts and relationships present in DOLCE. DOLCE's domain, i.e. the set of entities classified by the ontology's concepts (referred to as a set of *particulars*, PT) is divided into four sub-domains. For the purposes of this article, we consider only two of these here (see Figure 2):

- The *endurants* (ED). These are entities which "are in time" and are wholly present whenever they are present (objects, substances and also ideas). Within this sub-domain, one can distinguish physical objects (POB) and non-physical objects (NPOB), depending on whether or not the objects have a direct physical location.

- The *perdurants* (PD). These are entities which "occur in time" but are only partially present at any time they are present (events and states). Within this sub-domain, one can distinguish *events* (EV) and *statives* (STV) according to a "cumulativity" principle: the mereological sum of two instances of a *stative* (for example, "being seated") is an instance of the same type, which is not the case for the sum of instances of *events*, for example the K-CAP 2003 and K-CAP 2005 conferences. The latter (not being atomic) are considered to be *accomplishments* (ACC).

The main relationship between endurants and perdurants is that of *participation*, with PC(x,y,t) holding for: "x *(necessarily an endurant) participates in* y *(necessarily a perdurant) at time* t". For example, the co-authors of this article (endurants) participated in the *drafting* of the text (a perdurant).

More particularly in terms of the domain of knowledge which directly involves OntoKADS, DOLCE's commit-ment (corresponding to a consensus viewpoint within the AI and KE communities) can be summarized as follows[4]:

- Knowledge is the ability of an entity to perform an *action*, i.e. to produce changes in a world. The notion of "ability" implies that knowledge is of ideal order and that it does not coincide with any of the performed *actions*: knowledge is concerned with a mental world or, in other words, with *mental objects* (MOB) which are non-physical private objects for (which belong to) the entity, whereas the *action* is a perdurant.

- This knowledge or ability is embodied by an entity (the *agentive*[5]), which confers the latter with the potential to repeat *actions* in which it participates (in terms of the *PC* relationship) as an *agent*. Here, the *agentive* concept covers both the notion of an *intentional agent* [5] (i.e. an agent which is driven by a goal - representation of a desired world state) and that of a *rational agent* [19] (i.e. an agent which uses appropriate resources to achieve the goals that it has set itself).

---

[2] This ontology was chosen mainly for the reasons outlined in the Introduction but also because it integrates a class of mental objects which turn out to be very important for analyzing problem-solving knowledge.

[3] Even though the goals are different *a priori* (building an expertise model vs. verifying the logical coherence of subsumption links), we shall see that the labeling meta-properties are of the same nature.

The *action* (AC) is defined in DOLCE-Lite+ as an "*accomplishment exemplifying the intentionality of an agent*"[6]. The abstract characterization of *agent*/*agentive* applies to entities with very wide-ranging physical characteristics: a human being, a robot or a knowledge-based system.

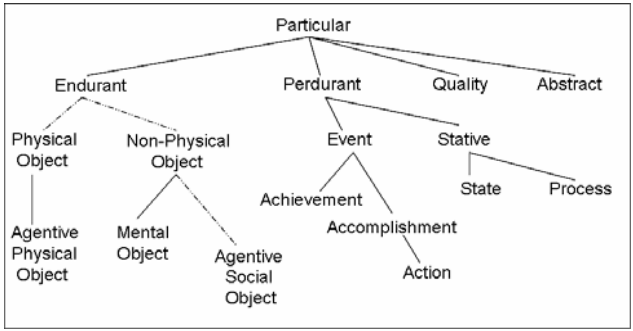

**Figure 2**. Excerpt from the DOLCE ontology

**The two main sub-ontologies in OntoKADS**

The OntoKADS ontology includes two sub-ontologies: a problem-solving ontology (which is independent of CommonKADS) and an ontology for modeling primitives close to those in CommonKADS. In the rest of this article and in order notably to illustrate the first sub-ontology, we will consider two examples of problem-solving situations: *diagnosing a car breakdown* and *calibrating a simulation code*. The first example corresponds to the teaching example dealt with in the CommonKADS reference book [21]. The second corresponds to an application which we are currently using to evaluate OntoKADS [4][7].

The first sub-ontology extends DOLCE in order to enable description of problem-solving activities. This proposed extension takes into account an analysis of existing problem-solving models and notably those created using the CommonKADS method. Analysis of the models shows that two distinct, general categories of *actions* suffice for tackling problem-solving activities, regardless of the latter's complexity: *Reasonings* and *Communications*[8].

We define *Reasonings* (e.g. diagnosing a breakdown, calibrating a simulation code, putting forward a hypothesis) as *actions* (A1) which aim at transforming the *agent-Reasoner's* mental world. *Reasonings* contrast in this respect with *actions*, which seek to transform the real world. The modification does not concern the real world but the *representation* that the *Reasoner* makes of the real world - in other words, *mental objects*. These are carried out by a human being or a (knowledge-based) system and may require the latter to interact with another human being or system - if only to exchange information with the outside world.

We liken an *Interaction* to an *action* whose performance is influenced by another *agentive*: this influence translates into participation (in one way or another) in the same *action* (D1). Within *Interactions*, we consider *Communications*, classified as *Interactions* whose goal is to exchange information[9] (A2). *Communications* may correspond to simple enunciations (e.g. presenting a result) or more complex *Interactions* made up of several enunciations involving different enunciators (e.g. obtaining/receiving information). On the basis of distinct goals (*transforming the reasoner's mental world* vs. *exchanging information*), we consider that *Reasonings* and *Communications* are distinct *actions* (A3).

(A1)     Reasoning(x) → AC(x)

(D1)     Interaction(x) $=_{def}$ AC(x) $\wedge \exists yzt$(isAgentOf(y,x) $\wedge$ z≠y $\wedge$ (APO(z) $\vee$ ASO(z)) $\wedge$ PC(z,x,t))

(A2)     Communication(x) → Interaction(x)

(A3)     Reasoning(x) → ¬Communication(x)

---

[4] More precisely, DOLCE's commitment appears to us to be coherent with this point of view.

[5] The *agentive* property in DOLCE is not found in isolation but corresponds to the extensional union of the *agentive physical object* (APO) and *agentive social object* (ASO) properties.

[6] DOLCE-Lite+ contains various DOLCE extensions under study. The extensions are presented in [17, chapt. 15].

[7] A simulation code implements a model of a class of systems. In order to simulate the behavior of a given system, it is necessary to calibrate the model by adjusting it to the particular characteristics of the system in question.

[8] In the rest of the article and in order to distinguish OntoKADS categories from those of DOLCE, the names of the former will be noted in a JAVA-like notation (e.g. *CalibrationData*, *isAuthorOf*), whereas abbreviations for the latter (e.g. ED, AC) will be used in the axioms.

[9] In the current version of OntoKADS, only these *Interactions* are considered. In particular, we do not take account of collaborative, problem-solving activities and the coordination mechanisms with which these activities are associated.

The second OntoKADS sub-ontology represents the modeling primitives of the CommonKADS method (or, more precisely, the new primitives defined for OntoKADS: *Task, Inference, FormalKnowledgeRole, DomainConcept*). Technically, we consider that this ontology is situated at a "meta" level with respect to the previous one, and thus corresponds to an ontology of *meta-properties*. These meta-properties enable classification of the problem-solving ontology's concepts according to the relation *CF(x,y,t)*, which stands for "x *classifies* y *at time* t"[10]. This relationship allows us to specify (for example) that the *Diagnosis* concept is modeled as a *Task* at a certain time $t_1$ (A4), that the *CalibrationData* concept is considered to be a *FormalKnowledgeRole* (A5) or indeed that *EmptyFuelTank* is a *DomainConcept* (A6).

The times $t_i$ correspond to building times for the expertise model, and the classifications can change over time: hence, according to the CommonKADS method, a particular *Reasoning* may be considered as an *Inference* at a given moment and as a *Task* at another moment (if combined with a decomposition method). The classifications are additionally accompanied by constraints which express the fact that a modeling primitive can only classify certain types of concepts in the problem-solving ontology. For example, (A7) and (A8) express the fact that the *Task* primitive can only classify concepts subsumed by the *Reasoning* concept and that the *TransferFunction* concept can only classify *Communications*[11]. Figure 3 provides a graphic summary of several constraints (the modeling primitives are noted in bold characters).

(A4)  classify(Task,Diagnosis,$t_1$)

(A5)  classify(FormalKnowledgeRole, CalibrationData,$t_2$)

(A6)  classify(DomainConcept,EmptyFuelTank,$t_3$)

(A7)  classify(Task,x,t) → subsumes(Reasoning,x)

(A8)  classify(TransferFunction,x,t) → subsumes(Communication,x)

In this section, we introduced OntoKADS' core problem-solving ontology by adopting the standpoint of the perdurants (*Reasonings* and *Communications*). In the next section, we tackle analysis of the endurants which participate in these perdurants. This leads us to define modeling primitives which correspond to knowledge roles in the CommonKADS method.

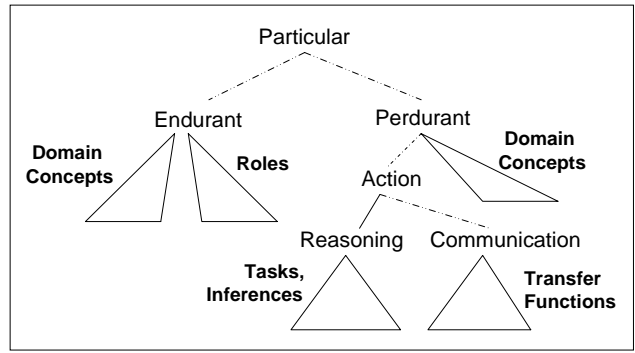

**Figure 3**. Structure of the OntoKADS ontology

## The OntoKADS kernel

In this section, we show how OntoKADS can answer the following questions: what is the ontological nature of the entities participating in *Reasonings* and *Communications*? How do these entities participate in *actions*? How can we characterize (in terms of meta-properties) the concepts representing such participation modes?

## Theoretical and practical knowledge and its objects

Let us return to the notion of knowledge. By admitting that all knowledge is knowledge about "something", about an "object", we can schematically distinguish between two categories of knowledge, depending on the nature of the objects (physical or mental) with which it deals [2]:

- *practical* knowledge (i.e. know-how "to act") deals with *physical* objects and enables action in the real world (e.g. banging in a nail, riding a bicycle).
- *theoretical* knowledge (i.e. know-how "to think") deals with *theoretical* objects (*mental objects*) and enables action in the mental world (e.g. calculating, deciding).

According to our definition, every *action* is based (at least in part) on theoretical knowledge - knowledge of the goal, representation of a desired world state. The goal certainly exists (as a mental object), whereas if the action fails, the desired world state may not be reached and thus may not exist. An *action* in the real world thus involves hybrid knowledge. What is the situation for *Reasonings* and *Communications*?

By defining *Reasoning* as an *action* seeking to transform the *agent's* mental world, we have forced its result - if one exists - to be a *mental object*. The term "result" (like the term "goal") can be likened to ways in which entities (*mental objects* in this instance) participate (as understood in DOLCE's *PC* relationship) in *Reasonings*. Two other participants enable us to define *Reasonings*: the *data* and the *solving method*. We complete the characterization of *Reasonings* by equally likening these participants to *mental objects*, which boils down to considering that by their very essence, *Reasonings* only involve theoretical knowledge[12]. As for *Communications*, and by complementing the participation of *mental objects*, the exchange of information between agents requires use of physical objects which "convey" information – "documents", in other words.

---

[10] This relation was introduced in [18] so as to account for the temporal classification of an instance by a concept. Here, we extend its signature by considering that the instance can be a concept classified by a meta-property.

[11] The relation *subsume(x,y)* signifies that all instances of the concept y are necessarily instances of the concept x.

In order to clarify the links between the *mental objects* participating in *Reasonings* and the documents on one hand and the way in which *mental objects* refer to a domain's physical objects on the other, it is necessary to specify their respective nature more fully. To achieve this, OntoKADS calls on the I&DA ontology, defined as an extension of DOLCE and designed to enable description of documents according to their content [8].

### Population of *mental objects* in I&DA

In order to account for documents and their content, I&DA distinguishes between three types of entity:

- The *Document*. A *Document* corresponds to a medium bearing a semiotic inscription of knowledge[13].

- The *Expression*. An *Expression* corresponds to the perceived signifier, expressed in a communication code. An example of an *Expression* is *Text* - a system of signifying units (words and phrases) structured according to language-defined rules.

- The *Content*. A *Content* corresponds to the *Expression's* signified. In functional terms, two signifieds can be distinguished: the *Proposition* (as a place of truth), and the *Concept*, (as a means of reference, i.e. for classifying entities). Let us add (and this a point that turns out to be important for OntoKADS) that a *Proposition* "uses" or "has for its subject" *Concepts* (the *hasForSubject* relation) (A9)(D2).

(A9)  hasForSubject(x,y) → Proposition(x) ∧ Concept(y)

(D2)  Subject(x) =$_{def}$ Concept(x) ∧ ∃y(Proposition(y) ∧ hasForSubject(y,x))

I&DA thus distinguishes between two general categories of *mental objects*, one (the *Expression*) depending on a communication code and the other (the *Content*) being code-independent. This enables us to consider that any given *Content* can be expressed (using different communication codes) by different *Expressions* and, equally, that any given *Expression* can be performed by different material inscriptions which may potentially solicit various senses on the receptor side (e.g. sound, visual or tactile inscriptions).

Finally, I&DA introduces a signifier-signified pairing which specializes in the communication of information between *agents*: the *Discourse* (likened to a statement) and its content, the *Message*. The *Message* is a *Proposition*, the result of a discursive act - for example, informing someone of something (for *Information*) or complaining about something (for a *Complaint*).

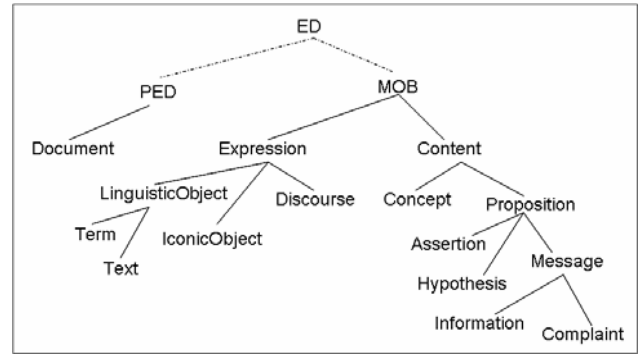

**Figure 4**. Top levels of the I&DA ontology

### Nature of the participating entities

The I&DA ontology enables one to specify the viewpoint adopted in OntoKADS in terms of the ontological nature of entities which participate notably[14] in *Reasonings*: these are *Contents* or, in other words, *Concepts* or *Propositions* which have *Concepts* as their *Subject*. This identity appears to be both pertinent and sufficiently general to account for the meaning of the following expressions (which appear as task data in CommonKADS expertise models): *EmptyFuelTank*, *EmptyFuelTankHypothesis*, *LowBatteryLevelComplaint* and *CarModel*.

This status of *Content* clarifies the link which exists between entities participating in *Reasonings* (which are *mental objects*), and the real world entities to which the *mental objects* refer: the real world entities only participate indirectly in *Reasonings*: that is to say, in as much as they are classified by a *Concept*. Hence, the expression "empty fuel tank" can be likened to the *Concept* of a class of states or to an *Assertion* (a *Proposition* considered to be true for an *agent*) whose *Subject* is the said *Concept*.

The usefulness of this interpretation is that one obtains a model which complies with the signature of DOLCE's *PC* participation relation: since a *Content* is an *endurant*, it is clear that an *endurant* participates in *perdurant-Reasoning*. It is also noteworthy that the *Concept* (either alone or playing the role of a *Subject*) can vary and may correspond to a *Concept* of state, process or behavior. The latter *perdurants* can then be classified according to an *agent's* perception mode (e.g. visible, invisible, observed, observable, etc.) or an abnormal (e.g. pathological) characteristic.

---

[12] This type of hypothesis may appear to be surprising, *a priori*. Hence, for example, a *Reasoning* such as a numerical calculation can be performed mentally, on paper, with a calculator or by using another agent-calculator. The point here is that the eventualities depend on the methods implemented in given calculation situations. The eventualities are therefore accidental and do not affect the essence of calculation. In contrast, we consider the *data* and the *solving method* to be essential participants.

[13] This definition enables one to account for very diverse documents, ranging from a sheet of paper bearing a form materialized by ink to the air around us carrying a sound wave which materializes what we hear.

[14] In the remainder of this section and for reasons of space, our main interest is *Reasonings* and, furthermore, we center our analysis on the *Reasonings'* data and results.

The above remark clarifies the generality of the modeling framework. In order to broaden its scope, we should add that a *Proposition* can be the result of particular *Reasonings* or indeed *Communications* when a *Message* is transmitted between agents. The expressions "empty fuel tank hypothesis" and "low battery level complaint" can thus be likened to *Propositions* resulting respectively from hypothetical *reasoning* and a discursive act consisting in "complaining about something".

In order to further emphasize the generality of this modeling framework, we note finally that a model (for example a *CarModel*) can be likened to a *Proposition*. This category covers knowledge models exploited by *Reasonings* as well as mathematical models used to simulate system behavior.

### Participation modes

However general it may be, the framework outlined so far remains incomplete because it does not allow us to account for expressions like "diagnosis hypothesis", "model to calibrate" or "calibrated model". This type of expression - useful for naming *knowledge roles* (in the CommonKADS sense) in task inputs and outputs [4] refers, in fact, to ways in which *Contents* participate in *Reasonings*, for example as data or results. This "participation mode" domain is covered by a specific OntoKADS component - a sub-ontology of "participation roles".

These roles (also referred to as "casual roles" or "thematic roles" in the literature) are defined in OntoKADS as particularizing the endurant concept. In fact, DOLCE's axiomization assimilates the notions of *endurant* and *participant*[15]. Hence, the participation roles or specialized participants are defined by introducing relationships which particularize the *PC* participation equation.

In this section and by way of illustration, we shall define first the *Patient* role using the *isAffectedBy* relation (A10)(D3)(T1) and then the specialized *Data* role using the *isDataOf* relation (A11-13)(D4)(T2-3). It is noteworthy that we have forced the *Data* i) to be a *Content* participating in an *Action* (A12) and ii) to participate from the start of the perdurant onwards (A13) (in contrast, the *Result* participates at the end of the perdurant). Finally, *CalibrationData* is defined as data for a particular *Reasoning* (a *Calibrating* (D5)) and a *ModelToCalibrate* is defined as a *Model* playing the role of *CalibrationData* (D6).

(A10)   isAffectedBy(x,y) $\rightarrow$ $\exists$t(PC(x,y,t))

(D3)    Patient(x) $=_{def}$ $\exists$y(isAffectedBy(x,y))

(T1)    Patient(x) $\rightarrow$ ED(x)

---

[15] According to DOLCE axioms: Ad33 (PC(x,y,t) $\rightarrow$ ED(x) $\wedge$ PD(y) $\wedge$ T(t)) and Ad35 (ED(x) $\rightarrow$ $\exists$y,t(PC(x,y,t))), only the endurants participate in the perdurants and, incidentally, all endurants participate necessarily in a perdurant.

(A11)   isDataOf(x,y) $\rightarrow$ isAffectedBy(x,y)

(A12)   isDataOf(x,y) $\rightarrow$ Content(x) $\wedge$ Action(y)

(A13)   isDataOf(x,y) $\rightarrow$ $\exists$t$\forall$t'((PRE(y,t') $\wedge$ t'$\leq$t)
              $\rightarrow$ PC(x,y,t'))

(D4)    Data(x) $=_{def}$ isDataOf(x,y)

(T2)    Data(x) $\rightarrow$ Patient(x)

(T3)    Data(x) $\rightarrow$ Content(x)

(D5)    CalibrationData(x) $=_{def}$ $\exists$y(isDataOf(x,y) $\wedge$
              Calibrating(y))

(D6)    ModelToCalibrate(x) $=_{def}$ Model(x) $\wedge$
              CalibrationData(x)

### The modeling primitive: *KnowledgeRole*

In our modeling of *Reasonings*, we were careful to characterize separately the nature of the participating entities on one hand and the nature of the participation modes on the other. In the CommonKADS method, this distinction reflects the difference between two modeling primitives, the "domain concept" primitive and the "knowledge role" primitive. In this section, we focus on the latter by showing how it can be ontologically founded. We end by defining novel modeling primitives for the OntoKADS method.

To achieve this, we have adopted the ontology of meta-properties defined in [12] as our reference framework. We also adopt their definition of three meta-properties involving a notion of "role": *role*, *formal role* and *material role*.

- A *role* is an *anti-rigid*[16] concept which depends on an external entity. Its *anti-rigidity*, (i.e. the property of being non-essential for all its instances) translates into dynamic behavior over time: an instance only plays a role by accident. Its *dependence* translates the fact that playing this role (for a given instance) necessarily implies the existence of another (external) instance.

- A *formal role* is a *role* which does not carry an *identity criterion*. A *formal role* restricts itself to characterizing a dependence mode vis-à-vis another entity, without constraining the identity of the entity playing the role. The *Agent* and *Patient* concepts (which we qualified as "participation roles") are examples of formal roles. At the beginning of this article, we notably saw that *agentives* possessing very varied *identity criteria* can play the role of *Agent*.

- A *material role* is a role carrying an *identity criterion*. A *material role* is usually subsumed by a *formal role* (from which it inherits its anti-rigidity and external dependence properties) and by a *type* (from which it inherits an *identity criterion)*. Examples of material roles are the *Student* and *Employee* concepts defined as *Per-*

---

[16] For reasons of space, we are not able to give the notions' formal definitions here. The reader is invited to refer to [12].

*sons* (a *type*) playing a formal role vis-à-vis a healthcare establishment or an employer.

By analogy with this reference framework and by particularizing it to the entities of the OntoKADS domain (i.e. entities participating in *Reasonings*) we define three notions of "reasoning roles" or "knowledge roles" (adopting the name of the primitive in CommonKADS). Figure 5 provides a graphical illustration of the concepts labeling using these meta-properties.

- a *KnowledgeRole* is a role which must be played by a *Content* and must depend on an *action*, a *Reasoning* or a *Communication*.

- a *FormalKnowledgeRole* is a *KnowledgeRole* which does not carry an identity criterion. The *Data* and *Result* concepts *(and more specifically CalibrationData and DiagnosisResult)* are examples of this. The roles can be played by *Concepts* or *Propositions* which are *Contents* carrying incompatible identity criteria.

- a *MaterialKnowledgeRole* is a *KnowledgeRole* carrying an identity criterion. The *ModelToCalibrate* and *DiagnosisHypothesis* concepts are examples of this. Each is subsumed (see Figure 5) by a *FormalKnowledgeRole* (*CalibrationData* and *DiagnosisResult* respectively) and by a type (*Model*, *Hypothesis*) which brings an identity criterion.

Finally, the resulting modeling framework for entities participating in *Reasonings* can be summarized as follows:

- The *KnowledgeRole*, *FormalKnowledgeRole* and *MaterialKnowledgeRole* modeling primitives are (like the other primitives) meta-properties, i.e. properties which classify other properties temporally.

- These meta-properties classify participation roles in *Reasonings*. The *Inputs* and the *Outputs* (primitives particularizing the *KnowledgeRole* primitive) classify *Data* and *Results,* respectively.

- The *Data* and *Results* are played by *Contents*, *Concepts* or *Propositions* which have *Concepts* as subjects.

- These latter *Concepts* classify the domain objects, their components and the states and processes in which these objects intervene.

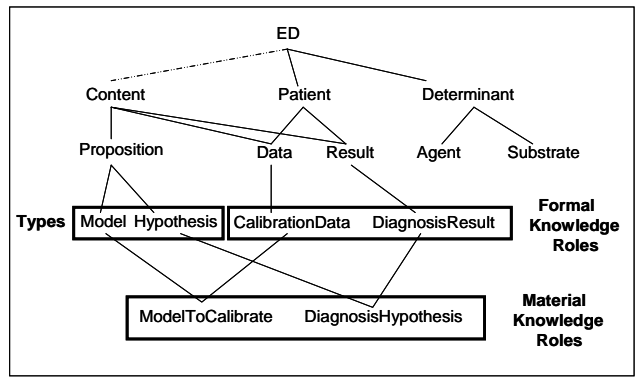

**Figure 5**. Labeling of participants using *KnowledgeRoles*

## DISCUSSION

The OntoKADS core problem-solving ontology prompts a knowledge modeling process which differs significantly from that proposed by CommonKADS. Schematically, OntoKADS requires more comprehensive specification of the ontological commitments for the conceptual entities concerned before "putting in primitives" these entities. The difference is most notable for *Reasonings* and their participants.

The OntoKADS modeling framework for *Reasonings* brings new light on the CommonKADS *knowledge role* primitive. The categories that are usually selected for knowledge roles in CommonKADS models correspond to categories of *Concepts* and *Propositions* in OntoKADS. These categories deal with the domain of representations: this explains why the terms designating knowledge roles may be independent of the field of application.

OntoKADS replaces the CommonKADS knowledge roles (which are poorly defined in semantic terms) by *MaterialKnowledgeRoles*, whose definition is based on a rigorous framework (in particular, these concepts must verify certain meta-properties). This framework provides the modeler with rules which allow him/her to identify these roles, by forcing him/her to answer the following questions: i) which type of representation is participating (e.g. message, model, assertion, hypothesis)? If applicable, what is the subject of the representation (e.g. a complaint about the state of the car, a model of the car's poor behavior, a hypothesis concerning the car's malfunction)?, ii) is this *Data* or a *Result*? and iii) what sort of *Reasoning* is concerned?

Application of these rules leads to expertise models which differ from those obtained using the CommonKADS method. Our current work involves comparing these models by using examples of generic tasks and methods from the literature.

## RELATED ONTOLOGICAL WORK

A critical aspect for OntoKADS is investigation of the theoretical objects which participate in *Reasonings* - objects which we have likened (using I&DA) to *Concepts* and *Propositions* [8] defined as *mental objects* (as understood

in DOLCE). Other ontological works have also tackled this field by pursuing a range of objectives:

- modeling document content for the SUMO ontology's "practical semiotics" [20] or, more specifically, the origin of information contained in web pages for Fox and Huang's ontology of *propositions* [9].
- modeling *reified* entities (e.g. standards, plans, methods, diagnoses, etc.) for the D&S ontology (Descriptions and Situations) [11].
- modeling an agent's *mental states* (e.g. beliefs, desires, intentions, etc.) and their links to *mental objects* for the COM ontology (Computational Ontology of Mind) [7].

The lack of a rigorous framework of definitions (such as that provided by the DOLCE functional ontology) in [20] and [9] entails that comparing conceptualizations is a delicate task. For example, entities referred to as "propositions" are defined by both ontologies as "abstract entities corresponding to document content". However, this characterization alone does not enable one to know (amongst other things) whether these are endurant or atemporal entities or whether the representations are subjective (agent-dependent) or objective. The two other ontologies (D&S and COM) are, in contrast, defined as extensions of DOLCE, which facilitates comparisons.

The COM ontology [7] appears to complement OntoKADS in all respects. Its domain is composed of two disjoint sub-domains: i) *mental states*, which one can consider as being parts (as understood in the *P* "is part of" relationship in DOLCE) of *Reasonings*. In particular, one can set any *Reasoning* (or, more generally, any *action*) to include an *intention*; and ii) *mental objects*, divided into *percepts* (direct representations of the real world) and *computed objects* (the results of cognitive processes). It appears clear that the *Concepts* and *Propositions* in OntoKADS can be defined as *computed objects*. On the other hand, one can set COM's *computed intentions* to correspond to *Propositions* in OntoKADS. On this basis, a merger between COM and OntoKADS seems to be possible.

The D&S ontology [11] and its extension to *information objects* (both of which are included in DOLCE-Lite+ [17, chapt. 15]) appear to cover the same domain as I&DA if one performs the following alignment: the *reified theories* or *s-descriptions* in D&S correspond to *Propositions* in I&DA, the *reified concepts* or *c-descriptions* correspond to *Concepts* and the *information objects* correspond to *Expressions*. In fact, the *s-descriptions* (following the example of *Propositions)* subsume entities as diverse as objectives, methods and diagnoses and are expressed by *information objects*. However, differences do exist, as shown by the recent application of D&S to construction of a core biomedical ontology [10].

Certain *c-descriptions* (referred to as *parameters*) reify constraints on *regions* (as understood in DOLCE). For example, according to D&S, the *fever* and *critical systolic blood pressure* parameters "select" a *body temperature* sub-region and a *blood pressure* sub-region, respectively. In contrast, in OntoKADS, such concepts are likened to concepts of states or processes (perdurants). These concepts can then play the role of *Signs* by enabling the evocation of other concepts [14]. Hence, the *Concept* in OntoKADS does not subsume (following the example of *c-description*) the *Fever* or *CriticalSystolicBloodPressure* concepts but does subsume the *Sign* concept, representing a role played by a *Concept* during a *Reasoning* (according to our definitions, the *Sign* concept is a *FormalKnowledgeRole*). The D&S *c-description* concept is thus quite different from the *Concept* in OntoKADS.

These comparisons demonstrate that ontological investigations in the field of knowledge objects are underway but that work is dispersed and, as we have seen, lacks consensus. Hence, at present, the domain is largely an open field for research. With OntoKADS, we are pursuing a dual objective: i) extending its domain so as to take account of other important knowledge objects for modeling problem-solving situations, notably *objectives* and *solving methods*; ii) challenging this modeling framework with the main generic tasks (*Reasonings*) and methods from the literature.

## CONCLUSION

In the present article, we have set the foundations of a radically ontology-centered approach to the construction of expertise models. Our OntoKADS method is based on a core problem-solving ontology labeled with meta-properties (modeling primitives) that the designer completes in order to account for application-specific reasonings. We defend the following thesis: recent progress in the field of formal ontologies (and notably work in the area of reasoning objects) means that such an approach is now within our reach.

Our work in defining OntoKADS continues in two directions. As we saw in the previous section, the ontology itself is being expanded. In addition, the design of a software environment for the method is now underway. This environment is defined as an extension of the TERMINAE ontology construction platform [1] and uses the OntoSpec semi-informal ontology specification language [15].